\documentclass[letterpaper]{article} % DO NOT CHANGE THIS
\usepackage{aaai25}  % DO NOT CHANGE THIS

\usepackage{times}  % DO NOT CHANGE THIS
\usepackage{helvet}  % DO NOT CHANGE THIS
\usepackage{courier}  % DO NOT CHANGE THIS
\usepackage[hyphens]{url}  % DO NOT CHANGE THIS
\usepackage{graphicx} % DO NOT CHANGE THIS
\usepackage{natbib}  % DO NOT CHANGE THIS AND DO NOT ADD ANY OPTIONS TO IT
\usepackage{caption} % DO NOT CHANGE THIS AND DO NOT ADD ANY OPTIONS TO IT
\usepackage{algorithm}
\usepackage{algorithmic}
\usepackage{multirow}
\usepackage{graphicx}
\usepackage{booktabs}
\usepackage{colortbl}
\usepackage{paralist}
\usepackage{enumitem}
\usepackage{amssymb}
\usepackage{bm}
\usepackage{makecell}
\usepackage{arydshln}
\usepackage{pifont}
\usepackage{longtable}
\usepackage{times}
\usepackage{latexsym}
\usepackage[T1]{fontenc}
\usepackage[utf8]{inputenc}
\usepackage{microtype}
\usepackage{inconsolata}

\usepackage{times}
\usepackage{latexsym}
\usepackage{multicol}
\usepackage{multirow}
\usepackage{subfigure}
\usepackage[T1]{fontenc}
\usepackage[utf8]{inputenc}
\usepackage{arydshln}

\usepackage{microtype}

\usepackage{inconsolata}
\usepackage{float}
\usepackage{graphicx}
\usepackage{subcaption}

\usepackage{newfloat}
\usepackage{listings}
\DeclareCaptionStyle{ruled}{labelfont=normalfont,labelsep=colon,strut=off} % DO NOT CHANGE THIS
\floatstyle{ruled}
\newfloat{listing}{tb}{lst}{}
\floatname{listing}{Listing}
\title{Explore the Potential of LLMs in Misinformation Detection: An Empirical Study}

\author {
    Mengyang Chen\textsuperscript{\rm 1,2},
    Lingwei Wei\textsuperscript{\rm 1,*},
    Han Cao \textsuperscript{\rm 1,2},
    Wei Zhou \textsuperscript{\rm 1},
    Songlin Hu \textsuperscript{\rm 1,2}
}
\affiliations {
    Institute of Information Engineering, Chinese Academy of Sciences \textsuperscript{\rm 1}Affiliation 1\\
    School of Cyber Security, University of Chinese Academy of Sciences \textsuperscript{\rm 2}Affiliation 2\\
    \{chenmengyang, weilingwei, caohan, zhouwei, husonglin\}@iie.ac.cn
}

\begin{document}

\maketitle

\begin{abstract}
Large Language Models (LLMs) have garnered significant attention for their powerful ability in natural language understanding and reasoning. In this paper, we present a comprehensive empirical study to explore the performance of LLMs on misinformation detection tasks. This study stands as the pioneering investigation into the understanding capabilities of multiple LLMs regarding both content and propagation across social media platforms. Our empirical studies on eight misinformation detection datasets show that LLM-based detectors can achieve comparable performance in text-based misinformation detection but exhibit notably constrained capabilities in comprehending propagation structure compared to existing models in propagation-based misinformation detection. Our experiments further demonstrate that LLMs exhibit great potential to enhance existing misinformation detection models. These findings highlight the potential ability of LLMs to detect misinformation.
\end{abstract}

\section{Introduction}

Misinformation    \cite{shu2017fake}
\footnote{Both fake news and rumors both belong to misinformation. For convenience of reference, in this article, we use the term "news" to refer to information.}
generally refers to false information that is spread deliberately to deceive people and cause severe negative impacts in the fields of national politics    \cite{fisher2016pizzagate}, economy    \cite{vosoughi2018spread}, and society    \cite{faris2017partisanship}. 
Existing misinformation detection methods mainly focus on news text and its propagation (engaged users, comments, retweet behaviors. etc). Text-based misinformation detection methods identify misinformation mainly by feeding extracted text features of the news into classifiers for classification   \cite{2011Information,ma2015detect,ma2016detecting,yu2017convolutional, luvembe2023dual, hamed2023fake}. 
% However, existing misinformation publishers often mimic the style of true news when composing misinformation. 
Consequently, some researchers incorporate the propagation of the news such as users    \cite{lu-li-2020-gcan,su2023mining} and propagation structure  \cite{2018Early,bian2020rumor,DBLP:conf/acl/WeiHZYH20,wu2023decor, chen2024propagation} into the detection process, achieving promising performance. 

% Due to the extensive pre-training, 
Large language models (LLMs) possess rich world knowledge and human-like preferences   \cite{bai2022training}, demonstrating comparable abilities to humans in many tasks   \cite{nori2023capabilities,bommarito2022gpt}. 
% In Text-based misinformation detection, although some preliminary studies have suggested that the detection performance of LLMs is slightly inferior to that of small language models (SLMs)\footnote{
% % The term "Small Language Models" does not have a formal definition yet. In this paper, we use 
% \textit{Small language models}"  refer to models with fewer parameters compared to LLMs, which include deep learning models and pre-trained models (such as BERT).}, LLMs can analyze the news content and generate weak supervision signals to assist existing models in making more accurate judgments   \cite{hu2024bad,huang2023harnessing,leite2023detecting}. 
In misinformation detection,  previous studies \cite{hu2024bad,huang2023harnessing,leite2023detecting} have shown that ChatGPT \cite{chatgpt} exhibits slightly lower performance than small language models (SLMs) 
\footnote{SLMs refer to models with smaller parameter counts compared to LLMs
\cite{hu2024bad}}
when using text content for misinformation detection.
% the LLM supporting ChatGPT \cite{chatgpt}, exhibits slightly lower performance than smaller language models when using text content for misinformation detection, according to previous studies \cite{hu2024bad,huang2023harnessing,leite2023detecting}.  
However, they only focus on specific LLMs and overlook the assessment of LLMs' ability to leverage social context, which is particularly crucial for misinformation detection. 
Therefore, a more comprehensive evaluation of LLMs in exploiting content and propagation is still under-explored for the task.

In this paper, an empirical study is first conducted to assess the ability of LLMs to detect misinformation.
% Different from previous studies   \cite{hu2024bad,huang2023harnessing,leite2023detecting},
We provide a more comprehensive analysis of LLMs on content and propagation-based misinformation detection. To explore the capabilities and potential of LLMs in misinformation detection, we divided the evaluation into two parts: an assessment of \textit{LLM-based} detectors and an assessment of \textit{LLM-enhanced} detectors.

In the evaluation of LLM-based detectors, we use different prompts to guide LLMs in misinformation detection directly. The results show that, under well-designed prompts, LLMs can achieve performance comparable to fine-tuned smaller models in text-based misinformation detection tasks. Different LLMs exhibit varying topic preferences and language preferences in tasks. In propagation-based misinformation detection, the performance of LLMs significantly lags behind existing detectors. This is because LLMs have difficulty understanding graph-structured propagation information through natural language alone.

In the evaluation of LLM-enhanced detectors, we utilize LLMs for data enhancement and feature enhancement, in combination with existing models for detection. The results indicate a significant improvement in detection performance. Specifically, LLM embeddings outperform the embeddings from mainstream text embedding models like BERT on most of the datasets, and their generative capabilities can simulate social user comments, enriching the propagation data available for detection.

The contributions of this work can be summarized as follows:

1) A comprehensive empirical study is conducted to evaluate LLMs in misinformation detection with diverse prompting methods. 
To the best of our knowledge, this is the first study to explore the understanding ability of multiple LLMs for both content and social context on social media.

2) We propose a novel evaluation framework that divides the analysis into LLM-based detectors and LLM-enhanced detectors, allowing for a deeper understanding of how LLMs perform independently and in combination with traditional models for misinformation detection.

3) Our study highlights LLMs' potential ability in misinformation detection, contributing to further exploration of LLMs' applicability in this task.

\section{Related Work}

\subsection{Misinformation Detection}

Misinformation detection aims to identify and judge the authenticity of news. 
Existing detection methods focus on the news text and its propagation on social media. 

Text-based methods involve extracting features from the news for classification. Early works rely on crafted features \cite{2011Information,ma2015detect,holan20162016}, which required significant manual effort and domain expertise. With the development of deep learning, many works have employed neural networks 
% such as RNN  \cite{ma2016detecting, 2019Detect, cheng2020vroc}, CNN  \cite{yu2017convolutional} and GNN \cite{vaibhav2019sentence}
to automatically learn high-level features  \cite{ma2016detecting, 2019Detect, cheng2020vroc,yu2017convolutional,vaibhav2019sentence}. 
% Some researches also incorporate additional information like stance  \cite{wu2019different} and emotions  \cite{giachanou2019leveraging}. 
% Driven by the need for richer language representations, 
Recent works employ pre-trained language models \cite{Devlin2019BERTPO,He2020DeBERTaDB,DBLP:conf/emnlp/0001HDZJMS22} to learn a better representation from content for detection \cite{kaliyar2021fakeBERT,jwa2019exbake,hamed2023fake,luvembe2023dual}.
% have been use in conjunction with deep learning models, a combination that could be more effective \cite{kaliyar2021fakeBERT,jwa2019exbake}.

% Incorporating social context, such as
propagation-based methods introduce user information  \cite{dong2018dual,nguyen2020fang,yuan2020early,DBLP:conf/sigir/DouSXYS21} and propagation structure \cite{Ma2017DetectRI,DBLP:conf/acl/WongGM18,bian2020rumor,song2021temporally,dou2021rumor,DBLP:conf/coling/Wei0L0H22,wu2023decor} to determine authenticity. 
% Some studies use bi-directional GCN \cite{bian2020rumor} to model propagation and dispersion. 
To solve the issues raised by incomplete and noise propagation, some works use the contrastive learning  \cite{10.1145/3511808.3557477,DBLP:conf/sigir/HeLZY21,sun2022rumor}, learned uncertainty \cite{DBLP:conf/acl/WeiHZYH20,wei2022modeling,chen2024propagation}, and propagation reconstruction \cite{DBLP:conf/coling/Wei00H22,wu2023decor} for robust structural representations. 
 
\subsection{Large Language Models}
Large language models (LLMs) refer to language models that are trained on large-scale text corpora and contain hundreds of billions (or more) parameters, such as GPT-3, PaLM, LLaMA, etc   \cite{zhao2023survey}. Researchers have found that "emerging capabilities" enable LLMs to significantly perform well while demonstrating special abilities that smaller models lack, such as contextual learning, instruction following, and step-by-step reasoning   \cite{fan2023bibliometric}, 
% Currently, the research and application of large language models involve multiple aspects, including pre-training techniques, adaptation fine-tuning techniques, usage techniques, capability evaluation, etc   \cite{naveed2023comprehensive}., and show broad application prospects in fields such as medicine, education, and finance, 
showing broad application prospects in many fields such as medicine, education, and finance \cite{naveed2023comprehensive}.

% LLMs possess rich world knowledge and strong reasoning capabilities, suggesting that they could potentially be use for misinformation detection  
LLM has received increasing attention in misinformation detection \cite{chen2024combating}.
Some works evaluate LLMs in using text content for misinformation detection  \cite{hu2024bad,huang2023harnessing,choi2023llms}.
They find that LLMs are not as good as existing text-based detection models and highlight the supporting role of LLMs in detecting misinformation, including professional analyses for news \cite{hu2024bad,huang2023harnessing}, different writing styles of LLMs \cite{wu2024fake} and propagation generation for misinformation\cite{nan2024let,wan2024dell}.
% For instance, LLMs can generate professional analyses for news, which can be use as weak supervision signals for subsequent detection tasks  \cite{hu2024bad,huang2023harnessing}. 
% Leveraging the exceptional writing capabilities of LLMs,   \citet{wu2023fake} rewrote news articles in different styles with ChatGPT, to train the model’s robustness against variations in news styles. 
Despite these advancements, the performance of multiple LLMs and their abilities to utilize propagation to detect misinformation remain unexplored.

\section{Preminarlies}

In this section, we present concepts, notations, and problem settings used in the work. 

\paragraph{Misinformation  Detection}
is to verify the authenticity of a given news article, we take misinformation detection as a binary classification problem, where each sample is annotated with a ground truth label indicating its authenticity. In text-based misinformation detection, each sample consists of a single news article so it can be seen as text classification. In propagation-based misinformation detection, each sample includes not only the news content but also user comments and the propagation structure formed between the news and the comments.

Formally, in propagation-based misinformation detection, Dataset $\mathcal{D}$ consists of $N$ samples and each sample is represented by $\mathcal{G} = (\mathcal{V,E})$, where $\mathcal{V} = \{r,c_1,...,c_N\}$ represents the features of the news $r$ and its comments ($c_1,...,c_N$),  $\mathcal{E}$ represents a set of explicit interactive behaviors, e.g., retweet.
The task objective of misinformation detection is to classify samples and determine whether the news is true (labeled as 1) or false (labeled as 0) by a binary classifier.

\paragraph{LLM-based Detectors}

\begin{figure}[t]
\centering

  \subfigure[LLM-based Detectors]{\includegraphics[width=0.85\linewidth]{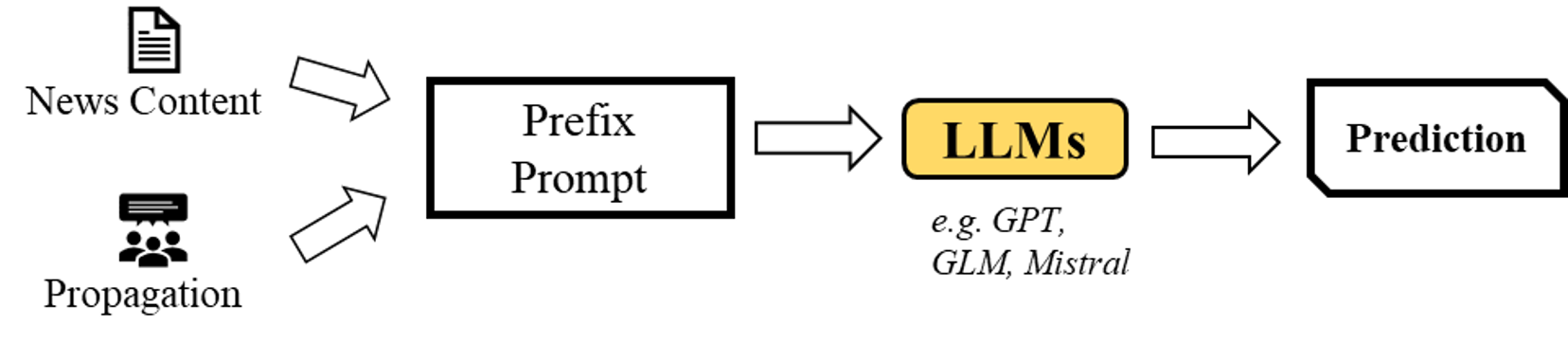}}\label{fig: LLM_pred}

  \subfigure[LLM-enhanced Detectors]{\includegraphics[width=0.85\linewidth]{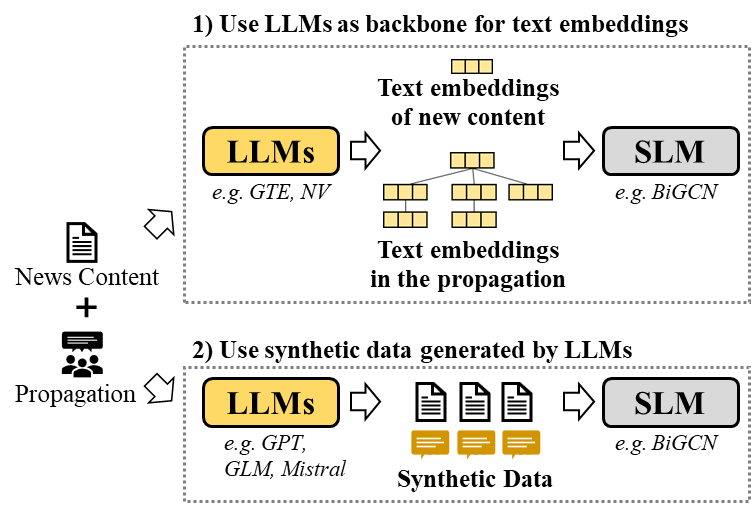}}
\caption{Two paradigms for utilizing LLMs in misinformation detection.}
  \label{fig:model}
\end{figure}

As shown in Figure 
\ref{fig:model}
(a), LLM-based detectors involve the direct utilization of large language models (LLMs) for misinformation detection. Specifically, we provide task-specific prompts to guide the LLMs in performing the detection task. 
\paragraph{LLM-enhanced Detectors}
LLM-enhanced detectors, on the other hand, use existing detection models as the core, with LLMs serving for data enhancement. As shown in Figure 1
\ref{fig:model}
(b). This is done in two forms: 1) Using LLMs as the backbone for text embeddings, SLMs utilize the embeddings generated by LLMs for detection. 2) Using synthetic data generated by LLMs as a supplement to existing data for SLMs.

\section{Evaluation of LLM-based Detectors}
In this section, we evaluate the performance of LLMs as detectors in misinformation detection. Specifically, we guide LLMs towards detecting misinformation through
various prompting methods and Instruction-Tuning.

\subsection{Evaluation Settings}
\paragraph{Datasets}
We experiment on eight representative misinformation detection benchmark datasets, consisting of three datasets containing news content and five datasets containing content and propagation. These datasets primarily come from news media and social platforms, covering topics such as politics, military, entertainment, society, and health. The statistics of datasets are listed in Table 
\ref{tab:datasets}
.
% Specifically, the datasets containing news contents are as follows.
\textbf{FakeNewsNet}   \cite{Shu2018FakeNewsNetAD} is a dataset compiled from GossipCop and PolitiFact, labeled as true or false by experts, primarily covering political and entertainment news from 2015 to 2017.
\textbf{LTCR}   \cite{Ma2023LTCRLC} is a Chinese rumor detection dataset with long sentences, sourced from official media during the COVID-19 pandemic. 
\textbf{FakeNewsDataset23} (FND23) is collected from fact-checking platforms such as Snopes and PolitiFact. It contains English news titles that were published after 2019, which are labeled as true or false.
% The datasets containing both news content and propagation are as follows.
\textbf{Twitter}, created by  \citet{siska-etal-2024-examining}, contains tweets published on Twitter\footnote{In July 2023, Twitter has been rebranded to X.}, and each tweet is annotated with true of false. 
\textbf{Twitter covid}   \cite{siska-etal-2024-examining} is a diverse COVID-19 healthcare misinformation dataset, including content and user social engagement.
% The data item for a tweet contains the tweet ID, text, corresponding label, user participation structure, etc.  
\textbf{PHEME5}    \cite{zubiaga2016learning} contains collections of rumors and non-rumors released on Twitter during 5 emergency events between 2014 and 2016.
% For each tweet, there are its reactions, annotation, and propagation structure composed of retweets.
 \textbf{CED}   \cite{song2018ced} contains Chinese rumor data scraped from Weibo, including forwarding and comment information related to the original Weibo posts, which are divided into rumors and non-rumors.
  \textbf{Weibo covid}   \cite{siska-etal-2024-examining} contains rumors about COVID-19 scraped from Weibo, which are divided into rumors and non-rumors.

Following the dataset split, we divide the datasets into training, validation, and test sets according to a ratio of 7:1:2. 

\subsection{Comparison Methods}\label{models}

\paragraph{Selection of SLMs}
For text-based detection models, following \cite{hu2024bad}, we selected 2 representative small models for text-based and propagation-based methods and 5 LLMs to evaluate LLM performance. we choose \textbf{Multi-Layer Perceptron (MLP)} and \textbf{EANN} \cite{wang2018eann} fine-tuned on the training set as comparison methods. For propagation-based detection models, we take Random Selection, Graph Convolutional Network (GCN), and a representative propagation-based detection method (i.e., \textbf{Bi-GCN}   \cite{bian2020rumor}) as baseline methods for comparison. 

\paragraph{Selection of LLMs}
For LLMs, we experiment with GPT-3.5-turbo, GLM, Mistral, Qwen and Vicuna for comparison. \textbf{GPT-3.5-turbo}
\protect\footnote{For detailed information about this model, please refer to https://platform.openai.com/docs/models/gpt-3-5}
is the most capable and cost-effective model in the GPT-3.5 family and in this experiment we use the default version \textit{gpt-3.5-turbo-0125}. 
\textbf{GLM}    \cite{Du2021GLMGL} is a general language model pre-trained with an autoregressive blank-filling objective and has been optimized for Chinese question-answering and dialogue. In the experiments, we use \textit{ChatGLM3-6B }
\footnote{https://huggingface.co/THUDM/chatglm3-6b}
. 
\textbf{Qwen} \cite{yang2024qwen2} is a language model series including decoder language models of different model sizes developed by Alibaba Cloud. In the experiments, we use \textit{Qwen1.5-7B-Chat}
\footnote{https://huggingface.co/Qwen/Qwen1.5-7B-Chat}
. 
\textbf{Mistral} \cite{jiang2023mistral}is the first dense model released by Mistral AI, designed to be ideal for experimentation, customization, and rapid iteration. We utilized \textit{Mistral-7B-Instruct-v0.2}
\footnote{https://huggingface.co/mistralai/Mistral-7B-Instruct-v0.2}
for the evaluation.
\textbf{Vicuna}  \cite{zheng2023judging} is a chat assistant trained by fine-tuning Llama 2 
% \cite{touvron2023llama}
on user-shared conversations collected from ShareGPT. The version we use in the evaluation is \textit{vicuna-7b-v1.5}
\footnote{https://huggingface.co/lmsys/vicuna-7b-v1.5}
.

\paragraph{Prompt Settings}

In this paragraph, we will clarify the prompt settings we use in evaluating the performance of LLMs as detectors in misinformation detection. As shown in Table 
\ref{tab:prompt_text}
and Table 
\ref{tab:prompt_propagation}
, to explore the potential of LLMs, and consider the critical role that prompt settings play in their task performance, we designed three \textbf{Vanilla prompts} (Shown in Table 
\ref{tab:prompt_text}
). Based on these Vanilla prompts, we developed additional prompts such as Task Prompt and Refine Prompt\footnote{Each type of these prompts also has three variants developed based on the Vanilla prompts, and we only present one variant of each type in Table 
\ref{tab:prompt_text}
and Table 
\ref{tab:prompt_propagation}
.}. 
% For detecting misinformation based on texts, we
\

For text-based detection, we utilize the following prompt settings:
\textbf{Task Prompt} \cite{yin2023did} involve incorporating task-related definitions into instructions to enhance the performance of LLMs on specific tasks. We introduce the definition of misinformation in vanilla prompts.
\textbf{Chain-of-Thought}   \cite{wei2023chainofthought} (CoT) prompt refers to a series of logically related thinking steps or ideas that are connected, forming a complete thinking process. We guide LLMs to carefully think through the detection process and then provide a judgment by adding "think step by step" in each prompt.
% \textbf{Few-shot Learning}: Few-shot Learning is a method where a small number of examples are added to the prompt to guide large models in performing tasks more effectively.
\textbf{Reason-aware Prompt} \cite{huang2024creation} helps LLMs assess misinformation by providing the underlying causes of misinformation within the prompt. 
Moreover, we employ Few-shot learning for LLMs, where the number of shots $N=\{0, 1, 2, 3, 4, 5, 6, 7, 8, 9\}$, evaluating its ability to generalize and adapt to detect misinformation.
In N-shot learning ($N>1$), the model is provided with $N$-shot exemplars from the training set, to aid its adaptation to the task. 

For propagation-based detection, we utilize the following prompt settings:
\textbf{Refine Prompt}: We enhance the vanilla prompts by instructing the LLMs to filter out comments unrelated to the news in the propagation information, reducing the impact of redundant information on the model.
\textbf{Format-Graph-Input (FGI)}: Inspired by the work of \cite{guo2023gpt4graph}, we describe the news propagation graph using formal graph terminology (e.g., nodes, edges) and then incorporate this into the vanilla prompts.

\paragraph{Instruction-Tuning}

For the Instruction-Tuning, we construct instruction-tuning datasets from the original datasets to fine-tune \textbf{LLaMa2-7B} \cite{touvron2023llama}, an open-source LLM developed by Meta AI with an upgraded size and performance for various NLP tasks. In the experiments, the temperature and top-p ratio are set to 0.6 and 0.9. 
 We fine-tune LLaMa2-7B by using the widely adopted and efficient Low-Rank Adaptation (LoRA) technique \cite{hu2021lora}. 

\paragraph{Evaluation Metrics.}
We run each prompt three times and report the average results. We use accuracy and macro-average F1 score as the evaluation metrics for
the evaluation. For the evaluation of each prompt of LLMs, we select the best result from its three variants.

For the details of the implementation, please refer to the Appendix.

% For details regarding the prompts use in the experiments, please refer to the appendix.

\subsection{Overall Performance on Misinformation Detection }
%基于文本的测评结果
\begin{table}[t]
    \centering
        \resizebox{1\linewidth}{!}{$
        \begin{tabular}{l|cc|cc|cc}
            \hline 
            \multicolumn{1}{c|}{\multirow{2}{*}{Methods}} & \multicolumn{2}{c|}{\multirow{1}{*}{FND23}} & \multicolumn{2}{c|}{FNN} & \multicolumn{2}{c}{LTCR}  \\ 
            & Accuracy & Macro-F1 & Accuracy & Macro-F1  & Accuracy & Macro-F1 \\ \hline 
            \multicolumn{1}{l|}{{\bf BERT}} &85.85 & 55.79&81.42 &\underline{71.34}& \textbf{95.63}&\underline{93.98} \\
            \multicolumn{1}{l|}{{\bf EANN}} & \textbf{89.23} & 47.34 &\underline{83.24} &\textbf{75.45} & \underline{95.38}& \textbf{94.22}\\
               \hline
            \multicolumn{1}{l|}{{\bf GPT3.5-Turbo}} & & & && & \\
            \multicolumn{1}{l|}{{\quad Vanilla prompt}} &82.56 & 55.62&58.87 &56.06&82.60&77.20  \\ 
            \multicolumn{1}{l|}{{\quad Task Prompt}} &87.50 &52.76 & 60.18&57.17& 83.77 &73.95 \\ 
             \multicolumn{1}{l|}{{\quad COT}} & 81.10&58.84 &73.98 &57.98&  81.47&65.07  \\ 
             \multicolumn{1}{l|}{{\quad RA Prompt}} &67.08 &51.74 & 68.29& 59.05& 80.64 & 73.99\\ \hline
            \multicolumn{1}{l|}{{\bf GLM}}  & & & && &  \\
            \multicolumn{1}{l|}{{\quad Vanilla prompt}} & 85.76&54.58 &64.76 &56.9& 86.87 & 79.59 \\ 
            \multicolumn{1}{l|}{{\quad Task Prompt}} &88.37 &46.91 & 58.72&53.84& 88.62 & 82.86\\ 
             \multicolumn{1}{l|}{{\quad COT}} & 87.79&46.75 &69.41 &60.04& 86.43 & 78.19 \\ 
             \multicolumn{1}{l|}{{\quad RA Prompt}} &85.47 &55.38 & 50.09& 48.06& 80.09 & 76.13\\  \hline 
            \multicolumn{1}{l|}{{\bf Qwen}} & & & && &  \\
            \multicolumn{1}{l|}{{\quad Vanilla prompt}} & 76.16&55.63 &65.14 &59.90& 90.81 & 85.83  \\ 
            \multicolumn{1}{l|}{{\quad Task Prompt}} &78.78 &58.32 & 67.17 &62.15&90.81 & 85.83 \\ 
             \multicolumn{1}{l|}{{\quad COT}} & 78.20&56.48 &82.82&49.28& 90.37& 85.49  \\ 
             \multicolumn{1}{l|}{{\quad RA Prompt}} &62.21 &50.00 & 67.77& 61.22& 90.37 & 84.55  \\ \hline
             \multicolumn{1}{l|}{{\bf Mistral}} & & & && & \\
            \multicolumn{1}{l|}{{\quad Vanilla prompt}} & 80.56&\textbf{61.74} &62.12 &60.70& 68.93 & 48.57  \\ 
            \multicolumn{1}{l|}{{\quad Task Prompt}} &83.24 &\underline{60.79} & 81.50&62.03& 75.68 & 47.97 \\ 
             \multicolumn{1}{l|}{{\quad COT}} & 86.81&53.57 &70.29 &63.03& 77.20 & 52.32  \\ 
             \multicolumn{1}{l|}{{\quad RA Prompt}} &83.33 &55.87 & \textbf{83.60}& 50.41&87.72  & 76.05 \\ \hline

             \multicolumn{1}{l|}{{\bf Vicuna}} & & & && & \\
            \multicolumn{1}{l|}{{\quad Vanilla prompt}} & 83.07&49.41 &70.42 &50.68& 75.17 &  49.72 \\ 
            \multicolumn{1}{l|}{{\quad Task Prompt}} &85.93 & 50.19 & 73.65&49.29& 76.09 & 51.09 \\ 
             \multicolumn{1}{l|}{{\quad COT}} & \underline{89.05} &47.10 &74.92 &44.04& 75.56 & 49.29  \\ 
             \multicolumn{1}{l|}{{\quad RA Prompt}} & 82.87 & 50.11 & 71.01 & 51.22 & 75.52 & 53.21 \\
                        \hline 

        \end{tabular}
$}
    \caption{Results on text-based misinformation detection. The best results for each dataset are highlighted in \textbf{bold}, and the second-best results are \underline{underlined}.}
    \label{tab:overall_results_content}
\end{table}

%基于传播的测评结果

\begin{table*}[t]
    \centering
\resizebox{0.9\linewidth}{!}{$
    \begin{tabular}{l|ccc|cc|cc|cc|cc|cc}
    \hline 
       \multicolumn{1}{c|}{\multirow{2}{*}{Methods}}          & \multicolumn{3}{c|}{\multirow{1}{*}{Inputs}}  &  \multicolumn{2}{c|}{\multirow{1}{*}{Twitter}} &\multicolumn{2}{c|}{\multirow{1}{*}{Twitter covid}} &\multicolumn{2}{c|}{\multirow{1}{*}{CED}}&\multicolumn{2}{c|}{\multirow{1}{*}{Weibo covid}} &\multicolumn{2}{c}{\multirow{1}{*}{PHEME5}} \\ 
       & News & Comments & Structure & Accuracy & Macro-F1  & Accuracy & Macro-F1  & Accuracy & Macro-F1  & Accuracy & Macro-F1  & Accuracy & Macro-F1  \\  \hline 
       \multicolumn{1}{l|}{\multirow{1}{*}{{Random}}} 
       & $\checkmark$ &  $\checkmark$ &   $\checkmark$ & 50.00 & 50.00 & 50.00 & 50.00 & 50.00  & 50.00 &  50.00  & 
 50.00 &  50.00  & 50\\ \hline            
       \multicolumn{1}{l|}{\multirow{1}{*}{{GCN$^\dag$}}} 
       & $\checkmark$ &  $\checkmark$ &   $\checkmark$ &77.20&76.61&76.67 & 75.08& 83.04  & 81.92 & 91.97  & 90.47
 & 80.31  &75.80  \\ \hline            
       \multicolumn{1}{l|}{\multirow{1}{*}{{Bi-GCN$^\dag$}}} 
       & $\checkmark$ &  $\checkmark$ &   $\checkmark$ &82.34&79.58&77.50 & 74.42 &93.45& 93.22&91.57 & 90.30 & 81.48& 79.71 \\ \hline   
       \multicolumn{1}{l|}{\multirow{2}{*}{{GPT-3.5-turbo}}} 
         & $\checkmark$ &  $\times$  &   $\times$     &    52.17 &  52.17  &   41.25&40.12& 48.28  & 48.28   & 43.37  &  41.19 &  54.80 &  51.59 \\ 
         & $\checkmark$ &  $\checkmark$ &   $\times$     &   57.33  &55.50 &  47.62  & 45.65&  65.36 &  65.26  & 37.18  & 35.90  & 46.33  & 46.10   \\ 
         \multicolumn{1}{l|}{{\quad Vanilla prompt}}& $\checkmark$ &  $\checkmark$ &   $\checkmark$  & 51.29  &51.18 & 56.36 &    \textbf{56.00}&  57.42 & 53.96   & 48.10  & 45.88  & \textbf{66.32}  &  48.44  \\ 
\hdashline
        \multicolumn{1}{l|}{{\quad Refine}} & $\checkmark$ &  $\checkmark$ &   $\checkmark$  & 56.58 &56.39 & \textbf{58.18} &    54.32&  60.60 &  58.63  &  \textbf{49.37} &  \textbf{49.16} & 63.49  &  49.28  \\ 
         \multicolumn{1}{l|}{{\quad FGI}}& $\checkmark$ &  $\checkmark$ &   $\checkmark$  & 56.71  &\textbf{56.64} & 56.64 &   53.75 & \textbf{74.77}  &  \textbf{74.66}  &  44.58 &  44.38 & 60.96  &  \textbf{57.61} \\ 
        \hline

       \multicolumn{1}{l|}{\multirow{2}{*}{{GLM}}} 
         & $\checkmark$ &  $\times$  &   $\times$     &    43.97 &  42.74  &   56.25 &55.06&  57.23 & 55.49  &  60.24 &  59.08 &  58.66 &  52.73 \\ 
         & $\checkmark$ &  $\checkmark$ &   $\times$     &   47.53  &44.00 &  \textbf{61.19}  & \textbf{59.15}&  \textbf{70.80} & \textbf{70.79} & \textbf{68.29}  & \textbf{68.12}  & 60.17  &\textbf{55.98}    \\ 
         \multicolumn{1}{l|}{{\quad Vanilla prompt}}& $\checkmark$ &  $\checkmark$ &   $\checkmark$  & 50.25  &46.15 & 54.72&    51.82 &  57.53 &  56.3  & 64.10  &  62.53 &  \textbf{68.52} &  47.16 \\ 
\hdashline
        \multicolumn{1}{l|}{{\quad Refine}} & $\checkmark$ &  $\checkmark$ &   $\checkmark$  & 52.43  &46.19 & 60.00 &   52.38 & 56.13  & 55.95  & 49.37  & 49.29  &  60.44 &  53.45 \\ 
         \multicolumn{1}{l|}{{\quad FGI}}& $\checkmark$ &  $\checkmark$ &   $\checkmark$  & \textbf{57.41}  &\textbf{57.04} & 48.57 &   46.43&  52.09 & 49.38   &  57.69 &  50.91 & 56.91  &  50.25 \\ 
        \hline 
       \multicolumn{1}{l|}{\multirow{2}{*}{{Qwen}}} 
         & $\checkmark$ &  $\times$  &   $\times$     &    59.05 & 49.94  &   57.50&43.33&  48.08 & 37.22   & 61.45  & 40.82  & 32.47  &  25.93 \\ 
         & $\checkmark$ &  $\checkmark$ &   $\times$     &   61.95  &\textbf{55.15} & 55.56  & \textbf{45.91} & \textbf{80.10}  & \textbf{80.05}   & \textbf{68.42}  & \textbf{56.32}  &  47.72 &  \textbf{47.71} \\ 
         \multicolumn{1}{l|}{{\quad Vanilla prompt}}& $\checkmark$ &  $\checkmark$ &   $\checkmark$  & 57.01  &37.29 & 52.63&   34.48 & 71.75  & 67.32   & 66.22  & 39.84  & 63.91  &  41.36 \\ 
\hdashline
        \multicolumn{1}{l|}{{\quad Refine}} & $\checkmark$ &  $\checkmark$ &   $\checkmark$  & 54.81  &36.35 & 55.26 &   40.35& 66.67   & 65.22  &  64.86 & 39.34  & 60.29  &  42.49 \\ 
         \multicolumn{1}{l|}{{\quad FGI}}& $\checkmark$ &  $\checkmark$ &   $\checkmark$  & 56.44  &37.88 &\textbf{58.75} &   37.01 & 56.19  &  36.67  & 66.27  & 39.86  &  \textbf{67.70} &  41.13 \\ 
        \hline 
       \multicolumn{1}{l|}{\multirow{2}{*}{{Mistral}}} 
         & $\checkmark$ &  $\times$  &   $\times$     &    56.00 &  54.56  &   50&49.97&   62.12 & 59.97   & 50.82  & 48.48  & 33.74  & 28.64   \\ 
         & $\checkmark$ &  $\checkmark$ &   $\times$     &   57.22   &55.27 &  41.38  & 29.27 & 65.03 & 64.36   & 50.00  & 49.43  &  \textbf{53.51} &  \textbf{52.77}   \\ 
         \multicolumn{1}{l|}{{\quad Vanilla prompt}}& $\checkmark$ &  $\checkmark$ &   $\checkmark$  & 48.84  &46.76 & \textbf{61.54} &    \textbf{60.61}&  \textbf{76.25} &  \textbf{74.35}  & 41.27  & 40.73  &  52.14 & 52.04   \\ 
\hdashline
        \multicolumn{1}{l|}{{\quad Refine}} & $\checkmark$ &  $\checkmark$ &   $\checkmark$  & 47.31 &43.56& 45.45 &   41.07 &  66.60  &  60.22  &  33.87 &  30.83 & 51.96  & 51.66 \\ 
         \multicolumn{1}{l|}{{\quad FGI}}& $\checkmark$ &  $\checkmark$ &   $\checkmark$  & \textbf{60.75}  &\textbf{60.47} & 44.64 &    44.63 &  74.77  &  74.66  & \textbf{54.55}  &  \textbf{54.49} &  41.31 &  38.00 \\ 
        \hline 
       \multicolumn{1}{l|}{\multirow{2}{*}{{Vicuna}}} 
         & $\checkmark$ &  $\times$  &   $\times$     &   53.68 &  38.06  &   48.75&48.74 & 53.86  &  53.65 & 56.63  &  \textbf{52.30}
 & \textbf{53.71}  &  \textbf{51.43} \\ 
         & $\checkmark$ &  $\checkmark$ &   $\times$     &  54.46   &\textbf{52.08} & 48.00  & 47.67&  56.20 & \textbf{55.28}  & \textbf{58.11}  &  50.51
 & 48.18  &  47.45  \\ 
         \multicolumn{1}{l|}{{\quad Vanilla prompt}}& $\checkmark$ &  $\checkmark$ &   $\checkmark$  & \textbf{55.33}  &47.76 &41.33 &   40.04 & 48.48  & 44.35  & 40.85  & 38.08 
 & 47.26  & 46.57  \\ 
\hdashline
        \multicolumn{1}{l|}{{\quad Refine}} & $\checkmark$ &  $\checkmark$ &   $\checkmark$  & 55.09  &51.73 & 39.29 &   38.02 &  \textbf{56.44} & 54.84  & 53.62  &  49.82
 &  45.03 &  44.82 \\ 
         \multicolumn{1}{l|}{{\quad FGI}}& $\checkmark$ &  $\checkmark$ &   $\checkmark$  & 49.51  &49.50 & \textbf{51.67} &    \textbf{51.00} &  46.03 &  46.02 & 54.24  &  51.51
 &  51.13 &  48.62 \\ 
        \hline 
\end{tabular}
$}
    \caption{Results on propagation-based misinformation detection. The best result for each method on each dataset is highlighted in \textbf{bold}. The \textbf{Structure} in Inputs means propagation structure of news.
    }
    \label{tab:overall_results_propagation}
\end{table*}

We evaluate the overall performance of text-based misinformation detection on FND23, FakeNewsNet, and LTCR, and that of propagation-based misinformation detection on the five datasets. The accuracy scores are presented in 
 Table 
 \ref{tab:overall_results_content}
 and Table 
 \ref{tab:overall_results_propagation}
 .
 % Table   \ref{tab:overall_results_content}  and Table   \ref{tab:overall_results_propagation}. 
 From the results, we can observe that:

\textbf{The performance of LLMs in text-based misinformation detection is impressive.} 
% As the accuracy of GPT-3.5-turbo is 6.8\% higher than that of fine-tuned DeBERTa on FakeNewsNet and 2.7\% lower on FND23, and the performance of GLM on LTCR is better than fine-tuned DeBERTa, this may be because LLMs can not learn semantic knowledge from the corpus like DeBERTa through fine-tuning. However, LLMs have learned real-world knowledge through large-scale pre-training, which enables them to make accurate judgments in detecting fake news by combining the background knowledge of the news. 
On the FND23 and FakeNewsNet datasets, LLMs achieve performance comparable to smaller models with specific prompts. For instance, on the FND23 dataset, Vicuna with the COT prompt achieves near-optimal accuracy, while Mistral with the RA prompt achieves optimal accuracy on the FNN dataset. This suggests that large models can leverage their extensive training data to make accurate judgments in fake news detection. However, LLMs still lag behind SLMs on the LTCR dataset. 
Additionally, we found that even though LLMs achieve a higher accuracy rate, the macro average F1 score remains low. This is due to the imbalance of positive and negative instances in the data (refer to Table 
\ref{tab:datasets}
), whereas SLMs eliminate this impact through training.

\textbf{Incorporating task-related information enhances the detection performance of LLMs}. From Table 
\ref{tab:overall_results_content}
, we can observe that when LLMs use task-specific prompts, their detection accuracy improves compared to vanilla prompts. This suggests that including a definition of fake news in the prompt can help LLMs better understand the task.

\textbf{Chain-of-Thought prompt and RA Prompt does not unequivocally enhance the performance of LLMs on misinformation detection.}
Some LLMs get reduced performance with the CoT Prompt compared to the Vanilla Prompt, showing difficulty for LLMs in reasoning tasks without extra cues. Like the COT Prompt, the RA Prompt failed to improve the performance of LLMs across all datasets equally. This reflects the limitations of current language models when dealing with multi-step reasoning tasks.

\textbf{LLMs possess certain but limited understanding abilities for propagation}.
In Table 2 \ref{tab:overall_results_propagation},  the detection performance of LLMs is better than Random.
However, the performance of prompt-tuning LLMs is significantly inferior to that of the small model (i.e., Bi-GCN). When only news text is provided, LLMs perform poorly in detection, indicating that additional information is needed for better detection. Although including comments and propagation structure can improve detection performance, the enhancement is often limited. Specifically, we find that when using a vanilla prompt, providing comment content to LLMs generally yields better results than additionally supplying propagation structure in most cases. Moreover, using different prompts to describe the propagation structure can significantly impact the performance of LLMs.

\textbf{The design of prompts has a significant impact on LLM's performance in detection.}
On two benchmarks, we find that different prompts have a substantial impact on LLM performance. For instance, on the FakeNewsNet dataset, using the RA prompt increases Mistral's detection accuracy by 21.48\% compared to using Vanilla Prompts. Similarly, in propagation-based benchmarks, prompts significantly affect how LLMs understand propagated information. For example, on the CED dataset, the accuracy of Mistral in detecting propagation-based misinformation differs by 9.65\% between using the Vanilla prompt and the Refine prompt. Furthermore, the way propagation information about the news is described (Vanilla and FGI) also affects how LLMs understand the propagated information. This indicates that careful design of prompts is essential when using LLMs for misinformation detection.

 \textbf{The performance of LLMs varies across different languages.} We find that Chinese LLMs (GLM and Qwen) perform better overall than the other three LLMs on Chinese datasets (LTCR, CED, and Weibo COVID). This may be because GLM and Qwen are specifically optimized for the Chinese language and context during their training, resulting in higher accuracy in handling Chinese misinformation.

\subsection{Further Analysis}

In this section, we explore the effects of Few-shot Learning and Instruction-Tuning on LLMs in text-based misinformation detection.

Then, through case analysis, we attempt to summarize the factors influencing LLMs in misinformation detection.

\begin{figure*}[htbp]
	\centering
  \subfigure[FND23]{\includegraphics[width=0.3\linewidth]{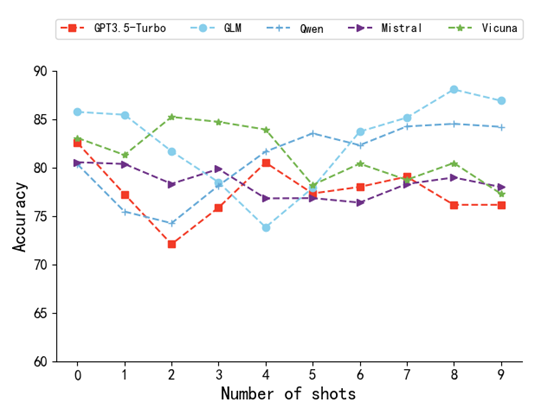}}
  \subfigure[FakeNewsNet]{\includegraphics[width=0.3\linewidth]{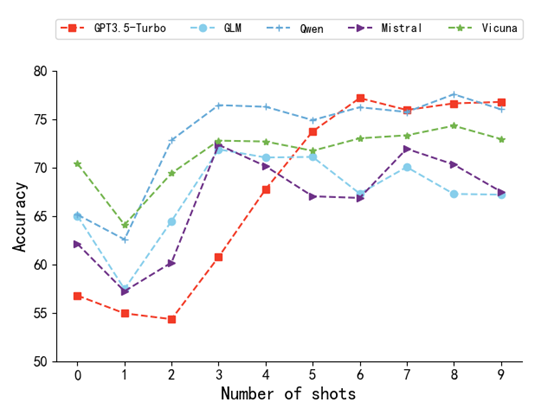}}
    \subfigure[LTCR]{\includegraphics[width=0.3\linewidth]{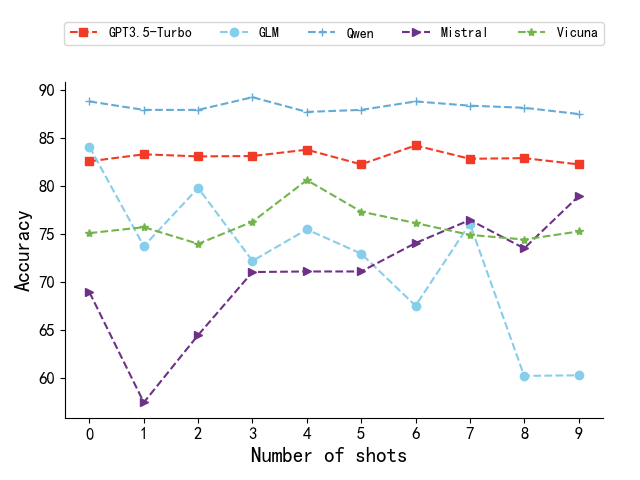}}

	\caption{Few-shot Learning results on three text-based misinformation detection datasets. }
	\label{fig:Few-shot}
\end{figure*}

\textbf{Few-shot learning does not consistently enhance the performance of LLMs}. As shown in Figure\ref{fig:Few-shot},
the performance of most LLMs (i.e., GPT-3.5-turbo and GLM) tends to decline with the initial few examples, it then gradually improves as the number of examples increases. 
Surprisingly, compared to the zero-shot setting, \textbf{most LLMs under one-shot learning show poor detection performance}. It indicates that when the number of training instances is insufficient, Few-shot learning fails to make the LLM  fully understand the misinformation detection task.

\textbf{Instruction-tuning could help LLMs in misinformation detection.} 
 The comparison of LLaMa2-7B's performance on eight datasets before and after instruction-tuning is shown in Table \ref{tab:llama_text}and Table\ref{tab:llama_propagation}. The results show that LLaMa2-7B gains significant performance improvements after instruction-tuning. However, the performance of the fine-tuned LLaMa is still limited compared to SLM, especially in propagation-based misinformation detection. Moreover, the fine-tuned LLaMa2-7B performs poorly on Twitter, Twitter covid, and Weibo covid with small data sizes, which may suggest challenges for instruction-tuning in low-resource scenarios.

\begin{figure}[t]
    \centering
    \includegraphics[width=0.7\columnwidth]{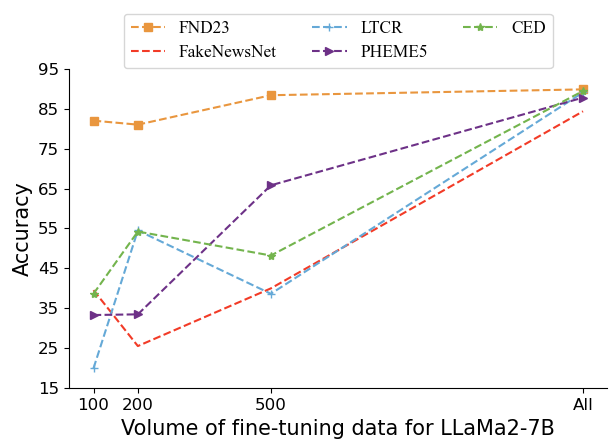}
    \caption{Accuracy scores (\%) of instruction-tuning LLaMa2-7B against different training data size.}
    \label{fig:tuning data size}
\end{figure}

To explore the relationship between the volume of fine-tuning data and the performance (accuracy) of LLaMA2-7B after instruction-tuning, we conducted experiments on all datasets except Twitter, Twitter covid, and Weibo covid due to their limited data volumes. Figure \ref{fig:tuning data size} illustrates that the performance of the fine-tuned LLaMA2-7B improves with an increase in training data volume, whereas insufficient data leads to suboptimal results.

\begin{table}[H]
    \centering
        \resizebox{0.8\linewidth}{!}{$
    \begin{tabular}{l|c|c|c}
    \hline 
       \multicolumn{1}{c|}{Models}          & FND23 & FakeNewsNet & LTCR  \\  \hline

       MLP$^\dag$              &   85.85   &81.42&   95.63  \\  \hline 

             EANN$^\dag$              &   89.23   &83.24&   95.38  \\  \hline 

       LLaMa2-7B     &      54.36 &	67.69&60.26
  
     \\ 
     LLaMa2-7B$^\dag$     &      89.83 &	84.35&89.08
  
     \\ 
        \hline 
\end{tabular}
$}
    \caption{Accuracy scores (\%) of LLaMa2-7B on text-based misinformation detection.
    % and report the best performance for each LLM.
    % We report the accuracy score of each method.
    $^\dag$ indicates the model is fine-tuned on the full training set.
    % For other LLMs, we report zero-shot detection performance.
    }
    \label{tab:llama_text}
\end{table}

\begin{table*}[htbp]
    \centering
\resizebox{0.75\linewidth}{!}{$
    \begin{tabular}{l|ccc|c|c|c|c|c}
    \hline 
       \multicolumn{1}{c|}{\multirow{2}{*}{Methods}}          & \multicolumn{3}{c|}{\multirow{1}{*}{Inputs}}  &  \multicolumn{1}{c|}{\multirow{2}{*}{Twitter}} &\multicolumn{1}{c|}{\multirow{2}{*}{Twitter coivd}} &\multicolumn{1}{c|}{\multirow{2}{*}{CED}}&\multicolumn{1}{c|}{\multirow{2}{*}{Weibo covid}}&\multicolumn{1}{c}{\multirow{2}{*}{PHEME5}} \\ 
       & News & Comments & Relations &  & & &
       % Zero-Shot &  Zero-Shot&  Zero-Shot&  Zero-Shot 
       \\  \hline 
       \multicolumn{1}{l|}{\multirow{1}{*}{{Random}}} 
       &- &-&-&50.00&50.00&50.00&50.00 &50.00\\ \hline
            \multicolumn{1}{l|}{\multirow{1}{*}{{GCN$^\dag$}}} 
       & $\checkmark$ &  $\checkmark$ &   $\checkmark$ &77.20&76.67&83.04&91.97 &80.31\\ \hline       
       \multicolumn{1}{l|}{\multirow{1}{*}{{Bi-GCN$^\dag$}}} 
       & $\checkmark$ &  $\checkmark$ &   $\checkmark$ &82.34&77.75&93.45&91.57&81.48 \\ \hline       
     
      \multicolumn{1}{l|}{\multirow{3}{*}{{LLaMa2-7B}}} 
        & $\checkmark$ &  $\times$  &   $\times$     &  43.54& 36.54 & 51.18  & 37.18 &48.34\\
        & $\checkmark$ &  $\checkmark$ &   $\times$   & 39.65 & 34.10    &   58.16&39.26&44.10 \\ 
        & $\checkmark$ &  $\checkmark$ &   $\checkmark$   &   34.58&  36.65& 54.23 &   35.32 &43.66   \\ 
        \hline 
      \multicolumn{1}{l|}{\multirow{3}{*}{{LLaMa2-7B$^\dag$}}} 
        & $\checkmark$ &  $\times$  &   $\times$     &  71.22& 66.55 &85.45   & 66.22&83.20 \\
        & $\checkmark$ &  $\checkmark$ &   $\times$   & 65.41 & 61.54   &   83.58&64.86&86.19\\ 
        & $\checkmark$ &  $\checkmark$ &   $\checkmark$   &   67.86&  58.24& 89.43 &   67.45&87.77    \\ 
        \hline 
\end{tabular}
$}
    \caption{Accuracy scores (\%) of LLaMa2-7B on propagation-based misinformation detection. We tune LLMs with vanilla prompts. 
    % and report the best performance for each LLM.
    % We report the accuracy score of each method.
    $^\dag$ indicates the model is fine-tuned on the full training set.
    % "\_" indicates the results that are worse than Random.
    }
    \label{tab:llama_propagation}
\end{table*}

\textbf{What affects LLMs in misinformation detection.}

Through analysis, We have currently identified four factors that may affect the detection capabilities of LLMs:

1) The limitation of input length: The input content to be detected exceeds the input length limitations of LLMs, resulting in the LLMs only being able to process a limited portion of the input. In our experiments, we find that LLMs often encounter errors during detection due to the length of the input tokens exceeding the model's maximum context length. As shown in Table \ref{tab:datasets}, the average token numbers of the Twitter covid dataset is approximately 78885, which has exceeded the maximum context length of Vicuna (4096 tokens). So we have to truncate the inputs to comply with the maximum context length of the LLMs;

2) Hallucination: The hallucinations of LLMs may affect their analysis of the content to be detected and lead them to make wrong judgments. We guide LLMs to generate judgments and analyses on the authenticity of the news by prompts. We then manually inspected the erroneous judgments and analyses, summarizing the following three reasons, as shown in Table \ref{tab:reasons}. According to \citet{ji2023towards}.'s definition, these three reasons can be categorized as illusions or misconceptions of LLMs. This is consistent with the conclusions from the study by \citet{hu2024bad};

\begin{figure}[H]
    \centering
    \includegraphics[width=0.5\columnwidth]{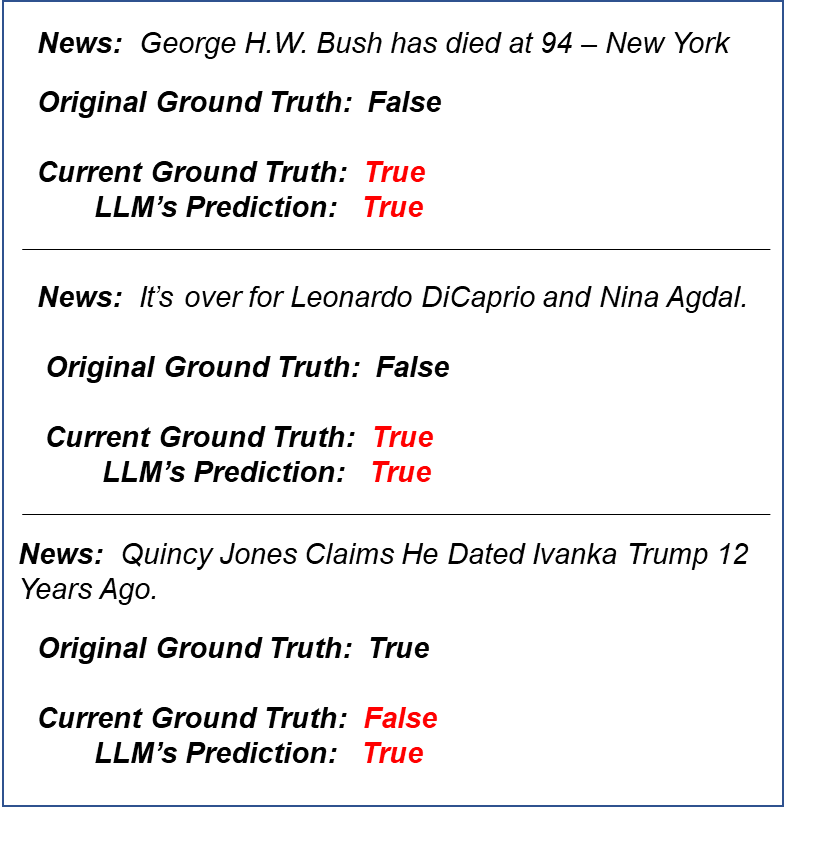}
    \caption{Changes in news authenticity lead to LLMs' misjudgment (Taking GPT-3.5-turbo as an example).}
    \label{fig:newstimelines}
\end{figure}

3) Variability in the authenticity of news: The changing authenticity of news over time may confuse LLMs. The accuracy of all LLMs decreases on the FakeNewsNet dataset compared with that on the FND23 dataset. We find that this discrepancy may be caused by the timeliness of the news through analysis of the samples misjudged by LLMs. As shown in Figure  \ref{fig:newstimelines}, the authenticity of some news has changed over time, which contributes to the decreased accuracy of LLMs in detecting misinformation. Moreover, SLMs essentially learn the mapping from text to labels during training, so the changes in the truthfulness of the news do not significantly impact their judgment;

4) Topic bias: We use \textit{GPT3.5-Turbo-0125} to perform topic classification on news from FND23, FNN, and LTCR, categorizing them mainly into a) political and military, b) technology and education, c) economics and finance, d) society, e) environment and health, f) entertainment. We explored the accuracy of LLMs in detecting misinformation across different categories of news, and the results are shown in Figure \ref{fig:topic}. It can be observed that LLMs also exhibit certain topic biases, they are more adept at judging news related to environmental and political topics compared to entertainment and social news.

\subsection{Dicussion}

This section primarily evaluates the ability of LLMs to act as predictors for misinformation detection through prompt-based experiments. The results show that in text-based misinformation detection, LLMs can perform comparably to or even surpass SLMs when given specific prompts. Including a definition of misinformation within the prompts helps LLMs better understand the task, thereby improving their performance. However, the prompt-based learning approach is not suitable for propagation-based fake news detection. This is because LLMs struggle to comprehend propagation information described in natural language, and propagation data, being a topological structure, is not well-suited to be input into the model as a sequence. We also find that LLMs are not well-suited for detecting outdated news, as the truthfulness of such news may change over time, affecting the LLMs' label predictions. Additionally, prompting LLMs to reason step-by-step or providing key indicators of fake news does not always enhance their performance. This is likely due to the inherent hallucinations within LLMs, which hinder their ability to correctly choose and summarize reasoning paths.

\section{Evaluation of LLM-enhanced Detectors}

In this section, we explore the potential of LLMs to enhance existing misinformation detection models on the previous tasks(text-based and propagation-based misinformation detection). We primarily consider utilizing LLMs for data enhancement on fake news datasets to improve the judgment capabilities of existing detection models. Specifically, there are two approaches for LLMs to enhance news data. One is feature-level enhancement, where LLMs leverage their extensive corpus knowledge to embed and enhance text features. The other approach is text-level enhancement, where LLMs use their generative abilities to produce new text, such as news analysis or commentary.

\subsection{Evaluation Settings}

\paragraph{Comparision Methods}
For feature enhancement using LLMs, we compare gte-Qwen2-1.5B-instruct as the encoder with BERT. gte-Qwen2-1.5B-instruct is a recently released General Text Embedding model designed to generate versatile and high-quality embeddings for various text-based tasks. We choose the SLMs in Section\ref{models} to evaluate the features generated by LLMs. Specifically, \textbf{w/ BERT} indicates that the text embeddings used by the model come from BERT. \textbf{w/NV} and \textbf{w/Qwen} indicates that the text embeddings come from \textit{NV-Embed-v2
\footnote{https://huggingface.co/nvidia/NV-Embed-v2}
}
and \textit{gte-Qwen2-1.5B-instruct
\footnote{https://huggingface.co/Alibaba-NLP/gte-Qwen2-1.5B-instruct}
}.
For text-level enhancement, we select ARG, GenFEND, and DELL as the comparison methods. \textbf{ARG} is a framework that combines large language models (LLMs) with small language models (SLMs), it integrates the reasoning information from LLMs to assist SLMs in making more accurate judgments in news analysis.
\textbf{GenFEND} is a generative feedback-enhanced fake news detection framework. It provides LLMs with diverse user profiles to generate comments from different groups and aggregates these generated comments for improved detection. We use BERT as the backbone of GenFEND. \textbf{DELL} is a fake news detection framework that integrates LLMs through a process involving generating news reactions to simulate diverse user perspectives, producing explanations to enrich context with sentiment and stance, predicting news with synthetic reactions using different expert Proxy Tasks, and merging expert predictions for a comprehensive judgment.

\paragraph{Implementation Details}

To eliminate the impact of language differences for a fair comparison, we use the \textit{bert-base-uncased} version of BERT for \textbf{w/ BERT} methods and the three text-level enhancement methods (ARG, GenFEND, and DELL) to get textual features on English datasets, and the \textit{bert-base-chinese} version to get textual features Chinese datasets\footnote{We also use the \textit{deberta-base} and \textit{deberta-base-chinese} for \textbf{w/DeBERTa} methods on English and Chinese dataset separately.}. To get the textual features from \textit{NV-Embed-v2} and \textit{gte-Qwen2-1.5B-instruct}, we set the task prompt as "\textit{Given a sentence, encode it into an embedding}."
For more details on the implementation of SLMs and the three text-level enhancement methods, please refer to Appendix \ref{implementation}. We use accuracy and macro-average F1 score as the evaluation metrics for
each model.

\subsection{Overall Performance of LLM-enhanced Detectors}

We evaluated the performance of LLMs as augmenters in previous tasks, and the results are shown in Table \ref{tab:Text-based detection_enhanced} and Table \ref{tab:Propagation-based detection_enhanced}. From the results, we have the following observations:

%基于文本的测评结果
\begin{table}[t]
    \centering
        \resizebox{1\linewidth}{!}{$
        \begin{tabular}{l|cc|cc|cc}
            \hline 
            \multicolumn{1}{c|}{\multirow{2}{*}{Methods}} & \multicolumn{2}{c|}{\multirow{1}{*}{FND23}} & \multicolumn{2}{c|}{FNN} & \multicolumn{2}{c}{LTCR}  \\ 
            & Accuracy & Macro-F1 & Accuracy & Macro-F1  & Accuracy & Macro-F1 \\ \hline 
            \multicolumn{1}{l|}{{\bf MLP}} & & & && & \\
            \multicolumn{1}{l|}{{\quad w/BERT}} &85.85 & 55.79&81.42 &71.34&95.63&93.98  \\ 
            \multicolumn{1}{l|}{{\quad w/DeBERTa}} &85.90 &52.76 & 82.24&72.50& 95.63 &93.98 \\ 
            \multicolumn{1}{l|}{{\quad w/NV}} &87.60 &66.01 & 85.04&79.05& 94.32 &82.05 \\ 
             \multicolumn{1}{l|}{{\quad w/Qwen}} & 86.05&62.75 &84.19 &77.28&  \textbf{100.00}&\textbf{100.00}  \\ 
 \hline
            \multicolumn{1}{l|}{{\bf EANN}}  & & & && &  \\
            \multicolumn{1}{l|}{{\quad w/BERT}} & 89.83&47.32 &83.04 &82.14& 95.63 & 93.562 \\ 
            \multicolumn{1}{l|}{{\quad w/DeBERTa}} &89.83 &47.32 & 84.24&83.84& 94.24 & 93.86\\ 
            \multicolumn{1}{l|}{{\quad w/NV}} &89.89&47.32 & \textbf{89.24}&\textbf{85.54}& 93.89&91.32 \\ 
             \multicolumn{1}{l|}{{\quad w/Qwen}} & 89.93&47.32 &88.56 &80.04& \textbf{100.00} & \textbf{100.00} \\ 
 \hline 
            \multicolumn{1}{l|}{{\bf ARG}} &\textbf{90.12} &50.17 &85.00& 77.52  &96.36 & 95.12 \\
            \multicolumn{1}{l|}{{\bf ARG-D}} & 89.83&\textbf{62.94} &84.35 &79.36& 96.36 & 95.12  \\ 
     
                        \hline 

        \end{tabular}
$}
    \caption{Results on text-based misinformation detection. The best results for each dataset are highlighted in \textbf{bold}.}
    \label{tab:Text-based detection_enhanced}
\end{table}

\begin{table*}[t]
    \centering
\resizebox{0.9\linewidth}{!}{$
        \begin{tabular}{l|cc|cc|cc|cc|cc}
            \hline 
            \multicolumn{1}{c|}{\multirow{2}{*}{Methods}} & \multicolumn{2}{c|}{\multirow{1}{*}{Twitter}} & \multicolumn{2}{c|}{Twitter covid} & \multicolumn{2}{c|}{CED}& \multicolumn{2}{c|}{Weibo covid}& \multicolumn{2}{c}{PHEME5}  \\ 
            & Accuracy & Macro-F1 & Accuracy & Macro-F1  & Accuracy & Macro-F1& Accuracy & Macro-F1& Accuracy & Macro-F1 \\ \hline 

            \multicolumn{1}{l|}{{\bf GCN}} & & & && & & &&& \\
            \multicolumn{1}{l|}{{\quad w/BERT}} &77.20 & 76.61&76.67 &75.08&83.04&81.92&91.97&90.47&80.31&75.80  \\ 
            \multicolumn{1}{l|}{{\quad w/DeBERTa}} &76.19 &75.74 & 75.17&75.65& 85.77 &83.95&90.32&89.78&82.02&79.14 \\ 
             \multicolumn{1}{l|}{{\quad w/NV}} & 78.87&75.84 &77.92 &75.58&  54.57&34.51 &91.17&88.42&81.48&78.67 \\ 
             \multicolumn{1}{l|}{{\quad w/Qwen}} & 78.87&75.84 &77.50 &74.42&  62.98&62.15 &\textbf{96.79}&\textbf{95.37}&82.00&79.51 \\ 
 \hline
            \multicolumn{1}{l|}{{\bf BiGCN}}  & & & && & & &&&\\
            \multicolumn{1}{l|}{{\quad w/BERT}} & 82.34&79.58 &77.50 &74.42& \textbf{93.45}& \textbf{93.22}&91.57&90.30&81.48&79.71\\ 
            \multicolumn{1}{l|}{{\quad w/DeBERTa}} &82.37 &80.91 & 78.72&77.84& 93.45 & 93.22&92.68&91.88&79.65&79.56\\ 
             \multicolumn{1}{l|}{{\quad w/NV}} & \textbf{89.18}&\textbf{88.98} &76.25 &75.38&  54.82  &   43.50 &91.578&90.07&\textbf{85.27}&\textbf{82.08} \\ 
             \multicolumn{1}{l|}{{\quad w/Qwen}} & 85.79&81.75 &75.00 &74.42&  61.21  &   61.12 &93.98&93.07&83.20&81.53 \\ 
 \hline 
            \multicolumn{1}{l|}{{\bf GenFEND}} &83.65 &80.88 &\textbf{80.20} &\textbf{78.86}& 88.23&86.01&95.48&94.66&83.03&69.79 \\
            \multicolumn{1}{l|}{{\bf DELL }} & 81.25&79.54 &76.65 &73.52& 90.79&89.87&25.67&91.26&82.02&82.21  \\

                        \hline 

        \end{tabular}
$}
    \caption{Results on propagation-based misinformation detection. The best results for each dataset are highlighted in \textbf{bold}.}
    \label{tab:Propagation-based detection_enhanced}
\end{table*}

1) \textbf{In most cases, feature enhancement based on LLMs can improve the detection performance of existing models.} On all three text datasets, the embeddings generated by LLMs improved the accuracy of MLP and EANN compared to those generated by BERT. In propagation-based misinformation detection, 
the embeddings generated by LLMs also led to performance improvements for GCN and BiGCN on datasets other than CED. This is because the larger scale and advanced pre-training strategies of LLMs enable their generated embeddings to more effectively capture and represent key features compared to embeddings generated by BERT and DeBERTa. Specifically, \textit{Qwen}, which was trained on a large amount of Chinese language data, produced better embeddings than \textit{NV} on the LTCR and Weibo COVID datasets. Conversely, on the English datasets Twitter and PHEME5, the situation was reversed.

2) \textbf{Despite the improvements observed in other datasets, the CED dataset and Twitter covid dataset presented an exception.}
When GCN and BiGCN use embeddings generated by LLMs on the CED dataset, performance declined significantly. Moreover, on the Twitter COVID dataset, GCN performed best with embeddings generated by NV, while BiGCN performed worst with embeddings generated by Qwen. This could be because the specific instructions use for generating embeddings with LLMs were not well-suited for embedding comments within the propagation graph on the datasets except for the reasons about language. Additionally, previous research \cite{purchase2022revisiting} has indicated that GNN models can show substantial performance variations based on the types of embeddings used, which might account for the decreased effectiveness in this case.

3) \textbf{The additional text information generated by LLMs is more effective when combined with the original data.} ARG and GenFEND outperform SLMs using BERT, while DELL does not. This is because both GenFEND and ARG integrate the generated data effectively with the original data through specially designed aggregation and interaction modules, whereas DELL only sets up different expert agents to make predictions based on the generated propagation graphs. We also noticed that the distilled version of ARG (ARG-D) achieved good results. This indicates that knowledge distillation can be use to leverage the benefits of LLM-generated information without the need to deploy the LLMs.

% This may be because ARG and GenFEND both fuse the embedded additional information generated by LLMs with the original text, allowing the small models to learn extra knowledge. In contrast, DELL generates propagation information based on the news text and uses this generated information for prediction, without leveraging the original propagation data. This could explain why DELL performs worse than ARG or even SLMs in some cases. We also noticed that the distilled version of ARG (ARG-D) achieved good results. This indicates that knowledge distillation can be use to leverage the benefits of LLM-generated information without the need to deploy the LLMs.
\subsection{Dicussion}
This section primarily evaluates LLM-enhanced detectors to explore the potential of LLMs as enhancers in existing misinformation detection methods. The experimental results show that LLMs when use as text embedding models, provide better embeddings compared to the mainstream embedding model BERT. However, the adaptability of Graph Neural Network (GNN) models to different types of embeddings may lead to scenarios where BERT outperforms LLMs. Additionally, using LLMs for text enhancement in datasets can provide richer information to the model, and combining LLM-generated analyses of original news data with the raw data produces unexpected results. These findings highlight the potential of LLMs in enhancing existing detection models.
\section{Conclusion}

We conduct an initial evaluation of LLMs in two approaches for misinformation detection: \textbf{LLM-based detectors} and \textbf{LLM-enhanced detectors}. The results show that LLMs can achieve comparable performance in text-based misinformation detection with well-designed prompts but exhibit notably constrained capabilities in comprehending propagation structures as a detector. On the other hand, LLM-enhanced detectors outperform SLMs in most cases when generating richer textual features, producing analysis, and simulating user engagements. Our findings demonstrate that LLMs hold great potential in detecting misinformation and supporting the existing misinformation detection models. In addition to the existing use of LLMs for synthesizing data to enhance text-level enhancement, our research also highlights the potential and prospects of utilizing LLMs as detectors and for feature enhancement in misinformation detection.

\section*{Limitations}

In this work, due to limitations in human resources and computational resources, we only evaluated LLMs at the 7B parameter level with limited prompts and have not yet assessed LLMs with larger parameter counts. Furthermore, while our study examined the performance of LLMs in detecting misinformation, it did not delve into whether there are other paradigms by which LLMs can enhance existing detection methods.

% \bibliography{aaai25}

\clearpage

\appendix

\section{Appendix}
\label{sec:appendix}

\begin{table*}[t]
    \centering
        \caption{The statistics of datasets. We tallied the number of samples in eight datasets, as well as the counts of positive and negative samples (real news and fake news). Additionally, we recorded the lengths of the news and comments within the samples, along with the number of tokens consumed when inputting them into LLMs\protect\footnote{We referenced the token count calculation from this URL: https://help.openai.com/en/articles/4936856-what-are-tokens-and-how-to-count-them}.     }

        \resizebox{1\linewidth}{!}{$
    \begin{tabular}{l|rrrrrrrr}
    \hline 
       Datasets        & \multicolumn{1}{c}{FND23}& \multicolumn{1}{c}{FakeNewsNet}& \multicolumn{1}{c}{LTCR} & \multicolumn{1}{c}{Twitter}  & \multicolumn{1}{c}{Twitter covid} & \multicolumn{1}{c}{CED}& \multicolumn{1}{c}{Weibo covid}& \multicolumn{1}{c}{PHEME}   \\ 
       \hline

       Number of News      &  1,720 &  23,196 &  2,287    &  1,154 &  400&  3,387 &411&5,801\\  
 
       Number of True News     & 347  &  17,441 &    1,678   &      575	& 252 &  1,538& 264 &3,829\\ 
     Number of False News    &  1373 &  5,755 &   609   &  579 &148 &1,849 &  147 &1972\\  

       Number of Comments   &  - &  - &  -    &59,255   &408,183 & 730,015& 36,779  &85,408 \\ 
       Max number of Comments     & -  & -  &   -   &674   &25,144 & 826& 3,097 & 674\\ 
       Average number of Comments    &  - & -  &   -   &52   &1,020 & 216& 90 & 15\\ 
              Max length of Comments(News)    & 394  & 340  &  4984    &146   &143 & 279& 697 & 187\\ 
       Average length of Comments(News)    & 106  & 69  &  207    &64  &58 & 12& 36 & 93\\ 
       Average token numbers per data entry    & 140  & 91  & 295   &4438   &78885 & 3456& 4322 & 1860\\ 

        \hline 
\end{tabular}
$}
    \label{tab:datasets}
\end{table*}

\begingroup
\begin{table*}[htb]
    \centering
    \small
    \vspace{2.8mm}
    \begin{tabular}{p{0.96\linewidth}}
    \toprule
    \textit{\ding{182}~Vanilla prompts} \\
    \vspace{-1mm}
    \textbf{1:} \textbf{Given the following news, we need you to detect fake news. Please determine whether each of the news is true or false. If it is true, please only answer 1, and if it is false, please only answer 0. Do not reply with any words other than 0 or 1. For each news, you must give your answer.} \\
    \textit{Given the news \texttt{[<news>]} , you are supposed to answer:} \\
    
    \textbf{2:} \textbf{Now you are an annotator to determine whether a given news is fake news. The news will be given in the form that 'Is [News] true news? Answer: [Answer]'. You need to give an answer in the [Answer] slot. There are two available answers that you can choose to fill the slot: 0 and 1, if the news is true, you are supposed to answer 1, otherwise, you should answer 0. You have to choose one of the two answers from 1 and 0. No need to answer any other words except 1 and 0.} \\
    \textit{Is \texttt{[<claim>]} a true statement? Answer: \texttt{[Answer]}} \\

    \textbf{3:} \textbf{Now we need you to detect fake news.  Below, I will provide you with a piece of news. If the news is true, please reply with 1. If the news is fake, please reply with 0. Do not reply any other words except 1 and 0.} \\
    \textit{News: \texttt{[<news>]}, Answer:} \\
    \midrule

\textit{\ding{183}~Task Prompt} \\
\vspace{-1mm}
 Now we need you to detect fake news. \textbf{\textit{Fake news is defined as news that is deliberately written to mislead readers into believing false information}}. Below, I will provide you with a piece of news. If the news is true, please reply with 1.If the news is fake, please reply with 0. Do not reply any other words except 1 and 0. 
\\ \midrule

\textit{\ding{184}~Chain-of-Thought (CoT) Prompt} \\
\vspace{-1mm}
Now we need you to detect fake news.  Below, I will provide you with a piece of news. If the news is true, please reply with 1.If the news is fake, please reply with 0. \textbf{\textit{Please consider the authenticity of the news step by step before replying}}. Do not reply any other words except 1 and 0.
\\ \midrule

\textit{\ding{185}~Reason-Aware (RA) Prompt} \\
\vspace{-1mm}
Now we need you to detect fake news. \textbf{\textit{Here are the options for categorizing the underlying causes of fake news: \
    [A. Emotional bias or misleading intent,\
    B. Lack of evidence or credible sources,\
    C. Conflicting fact,\
    D. Informal statements, expressions, or vague language,\
    E. insufficient supporting materials,\
    F. Lack of context or taken out of context,\
    G. Misinterpretation or misquotation,\
    H. Oversimplification or exaggeration,\
    I. Doctored images or videos,\
    J. Other reasons]\
    }}
     Below, I will provide you with a piece of news. If the news is true, please reply with 1.If the news is fake, please reply with 0. Do not reply any other words except 1 and 0."
\\ \midrule

\textit{\ding{185}~Few-shot Learning (N=3)} \\
\vspace{-1mm}
Now we need you to detect fake news. Below, I will provide you with a piece of news. If the news is true, please reply with 1. If the news is false, please reply with 0. Do not reply with any other words except 1 and 0.\\
\textbf{\textit{Here are some examples of fake news examples:}} \\
\textbf{\textit{News: [No Evidence Lisa Loring's Cause of Death Was COVID-19 Vaccine], Answer: [0]}} \\
\textbf{\textit{News: [AP: Diamond's Cause of Death Was Heart Disease, Not COVID-19 Vaccine], Answer: [0]}} \\
\textbf{\textit{News: [The U.S. Department of Defense awarded a contract for COVID-19 research in Ukraine months before the virus was known.], Answer: [0]}} \\
\textbf{\textit{Here are some examples of true news examples:}} \\
\textbf{\textit{News: [A photo shows Ukrainian soldiers posing in front of a NATO flag, an Azov flag, and the swastika-bearing flag once use by the Hitler Youth], Answer: [1]}} \\
\textbf{\textit{News: [A video authentically shows a March 2023 incident involving a U.S. surveillance drone and Russian Su-27 fighter jets], Answer: [1]}} \\
\textbf{\textit{News: [HIV is contagious.], Answer: [1]}} \\

\bottomrule
    \end{tabular}
    \caption{ Prompt settings use to evaluate the performance of LLMs as detectors on text-based misinformation. For vanilla prompts, Template prompts are in \textbf{bold}, the query is in \textit{italics}, and the input (the news to detect) is in \texttt{typewriter font}. For other prompts (such as Task Prompt), the changes compared to the Vanilla prompts are highlighted in \textbf{bold}.}
    \label{tab:prompt_text}
\end{table*}
\endgroup

\begingroup
\begin{table*}[htb]
    \centering
    \small
    \vspace{2.8mm}
    \begin{tabular}{p{0.96\linewidth}}
    \toprule
    \textit{\ding{182}~Vanilla prompts} \\
    \vspace{-1mm}
    \textbf{News:} \textbf{Now you are an annotator to detect misinformation. You need to determine whether the given claim is misinformation. The claim will be given in the format \texttt{[Claim[id:claim]]}} \\
    \textit{Given the claim \texttt{[Claim [<id, claim>]]}, please determine whether the claim is misinformation based on its comments and propagation structure. If the claim is true, you are supposed to answer 1; if it is false, you are supposed to answer 0. You have to choose one of the two answers, 0 or 1. No need to answer any other words except 0 or 1. Your answer is:} \\
    \vspace{2mm}
    \textbf{Comments:} \textbf{Now you are an annotator to detect misinformation. You need to determine whether the given claim with comments is misinformation. The claim and its comments will be given in the format \texttt{[Claim[id:claim]]}, \texttt{[Comments [comment\_texts]]}} \\
    \textit{Given the claim \texttt{[Claim [<id, claim>]]}, \texttt{[Comments [<id, comments>]]}, please determine whether the claim is misinformation based on its comments. If the claim is true, you are supposed to answer 1; if it is false, you are supposed to answer 0. You have to choose one of the two answers, 0 or 1. No need to answer any other words except 0 or 1. Your answer is:} \\
\\
    \textbf{Propagation Structure:} \textbf{Now you are an annotator to detect misinformation. You need to determine whether the given claim with comments and propagation structure composed of comments is misinformation. The claim, its comments, and the propagation structure composed of comments will be given in the format \texttt{[Claim[id:claim]]}, \texttt{[Comments [id:comment\_texts]]}, \texttt{[propagation structure: ['id1 replied to id2',...]]}. \texttt{[id:comment\_text]} means the ID and text of the comments on the tweet, and \texttt{'id1 replied to id2'} means the comment with ID \texttt{id1} is a reply to the comment with ID \texttt{id2}.} \\
    \textit{Given the claim \texttt{[Claim [<id, claim>]]}, \texttt{[Comments [<id, comments>]]}, and \texttt{[propagation structure [<propagation structure>]]}, please determine whether the claim is misinformation based on its comments and propagation structure. If the claim is true, you are supposed to answer 1; if it is false, you are supposed to answer 0. You have to choose one of the two answers, 0 or 1. No need to answer any other words except 0 or 1. Your answer is:} \\
    \midrule

\textit{\ding{183}~Refine Prompt} \\
\vspace{-1mm}
Now you are an annotator to detect misinformation. You need to determine whether the given claim with comments and propagation structure composed of comments is misinformation. The claim, its comments, and the propagation structure composed of comments will be given in the format [Claim[id: claim]], [Comments [id:comment\_texts]],[propagation structure: ['id1 replied to id2',...]]. [id:comment\_text] means the ID and text of the comments on the tweet, and 'id1 replied to id2' means the comment with ID id1 is a reply to the comment with ID id2.\\
 \textbf{Due to the noise in the comments and dissemination structure of the tweet, before you make a judgment, you need to remove comments that are irrelevant to the authenticity of the tweet and optimize the propagation structure based on the remaining replies, and then judge whether the tweet is a rumor based on the propagation structure optimized by yourself.}
\\ \midrule

    \textit{\ding{184}~Format Graph Input} \\
    \vspace{-1mm}
    Now you are an annotator to detect rumors. You need to determine whether the given claim is misinformation based on its propagation graph. The claim will be given in the form \texttt{[ID: claim]} and its propagation graph will be given in the form: \\
    \textbf{\texttt{Nodes: ['id1:content1', 'id2:content2', ...]}} \\
    \textbf{\texttt{Edges: [[All parent nodes' IDs], [All child nodes' IDs]]}} \\
    \textbf{\texttt{Root node: ['id of the root node: the content of the root node']}} \\

\bottomrule
    \end{tabular}
    \caption{Prompt settings use to evaluate the performance of LLMs as detectors on propagation-based misinformation. We only present one Vanilla prompt and its variants.}
    \label{tab:prompt_propagation}
\end{table*}
\endgroup

\subsection{Implementation Details}\label{implementation}
This section mainly explains the configurations of the models use in the evaluation.

For text-based methods, the learning rate for MLP (2-layers) and EANN is set to 2e-5, with a batch size of 32, and each training session runs for 20 epochs. We implement ARG
\footnote{https://github.com/ICTMCG/ARG}
and ARG-D under the same parameter setting reported in the original papers.

For propagation-based methods, the learning rate for GCN (2-layers) and BiGCN is set to 1e-3, with a batch size of 128, and each training session runs for 35 epochs. We implement GenFEND
\footnote{https://github.com/ICTMCG/GenFEND}
and DELL
\footnote{https://github.com/whr000001/DELL}
with \textit{Mistral-7B-Instruct-v0.2} on English datasets and \textit{Qwen1.5-7B-Instruct} on Chinese datasets, and we use the same parameter setting reported in the original papers. And for the proxy tasks in DELL, we removed the \textbf{Knowledge Retrieval} task.

The embedding models use by SLMs (MLP, EANN, GCN, and BiGCN) and LLM-enhanced methods (ARG, GenFEND and DELL) are shown in Table \ref{tab:embedding setting}.  For methods using BERT and DeBERTa, the Chinese versions (bert-base-chinese, deberta-base-chinese 
\footnote{https://huggingface.co/KoichiYasuoka/deberta-base-chinese}
) are use for Chinese datasets, while the English versions (bert-base-uncased, deberta-base 
\footnote{https://huggingface.co/microsoft/deberta-base}
) are use for English datasets.

All experiments were conducted on a single Tesla V100 (32GB), using PyTorch version 1.12.1 and Geometric version 2.3.1.

\begin{table*}[t]
    \centering
    \small
    \setlength{\tabcolsep}{3.5pt}
    \begin{tabular}{@{}clc@{}}
    \toprule
    \textbf{Methods} & \textbf{Text embedding methods} & \textbf{Feature dimension} \\
    \midrule
        \multirow{2}{*}{SLMs(SLMs w/BERT) \& Text-level enhancement methods} & bert-base-uncased  &768 \\
       & bert-base-chinese  & 768 \\

        \midrule
            \multirow{2}{*}{SLMs w/DeBERTa} & deberta-base  & 768  \\
       & deberta-base-chinese  & 768 \\

        \midrule
        \multirow{1}{*}{SLMs w/NV} & NV-Embed-v2\footnote{https://huggingface.co/nvidia/NV-Embed-v2}  & 4096  \\
        \midrule
        \multirow{1}{*}{SLMs w/Qwen} & gte-Qwen2-1.5B-instruct\footnote{https://huggingface.co/Alibaba-NLP/gte-Qwen2-1.5B-instruct}  & 1536  \\

     \bottomrule
    \end{tabular}
    \caption{Text embedding models use by different methods.}
    \label{tab:embedding setting}
\end{table*}

\begingroup
\begin{table*}[htbp]
    \centering
    \small
    \vspace{2.8mm}
    \begin{tabular}{p{0.96\linewidth}}
    \toprule
    \textit{\ding{182}~Failure to effectively integrate analysis results} \\
    \vspace{-1mm}
    \textbf{Q:} \textbf{Now we need you to detect fake news.  Below, I will provide you with a piece of news. If the news is true, please reply with 1. If the news is fake, please reply with 0. \textbf{\textit{Please consider the authenticity of the news and provide your reasons along with your answer.}}. } \\
    \textit{News: \textit{[<Joe Bidden closed schools during the COVID-19 pandemic>]}, Answer:} \\
    \textbf{A:} Joe Bidden did not close schools during the COVID-19 pandemic. Schools were closed as a result of government-issued shutdown orders to migrate the spread of the virus. Answer: \textbf{True}.\\
    \textbf{Label: Fake}\\
    \midrule
\textit{\ding{183}~Wrong analytical process} \\
\vspace{-1mm}
    \textbf{Q:} \textbf{Now we need you to detect fake news.  Below, I will provide you with a piece of news. If the news is true, please reply with 1. If the news is fake, please reply with 0. \textbf{\textit{Please consider the authenticity of the news and provide your reasons along with your answer.}}. } \\
    \textit{News: \textit{[<COVID Survivor Claire Bridges Sets Record Straight on Misleading Tweet About Her>]}, Answer:} \\
    \textbf{A:} Since the news is focuse on correcting information and suggests that there was a tweet circulating that contained inaccurate information about Claire Bridges, she has since clarified the facts to correct the record, which makes the statement true. Answer: \textbf{True}.\\
    \textbf{Label: Fake}
\\ \midrule

    \textit{\ding{184}~Unrelated analyses} \\
    \vspace{-1mm}
    \textbf{Q:} \textbf{Now we need you to detect fake news.  Below, I will provide you with a piece of news. If the news is true, please reply with 1. If the news is fake, please reply with 0. \textbf{\textit{Please consider the authenticity of the news and provide your reasons along with your answer.}}. } \\
    \textit{News: \textit{[<A music video features Ukrainian President Volodymyr Zelenskyy dancing in high heels, a cropped top, and skin-tight pants with three other men.>]}, Answer:} \\
    \textbf{A:} This statement is about Zelenskyy dancing in high heels, not a newsworthy event. This may be a clickbait headline and seems to be a fabrication. Answer: \textbf{Fake}.\\
    \textbf{Label: True}\\
\bottomrule
    \end{tabular}
    \caption{Three main factors contributing to worse
performance of LLMs: (a) failure to effectively integrate analysis results, (b) wrong analytical process, and (c) unrelated analyses}
    \label{tab:reasons}
\end{table*}
\endgroup

\begin{figure*}[htbp]
\centering

  \subfigure[ChatGPT]{\includegraphics[width=0.32\linewidth]{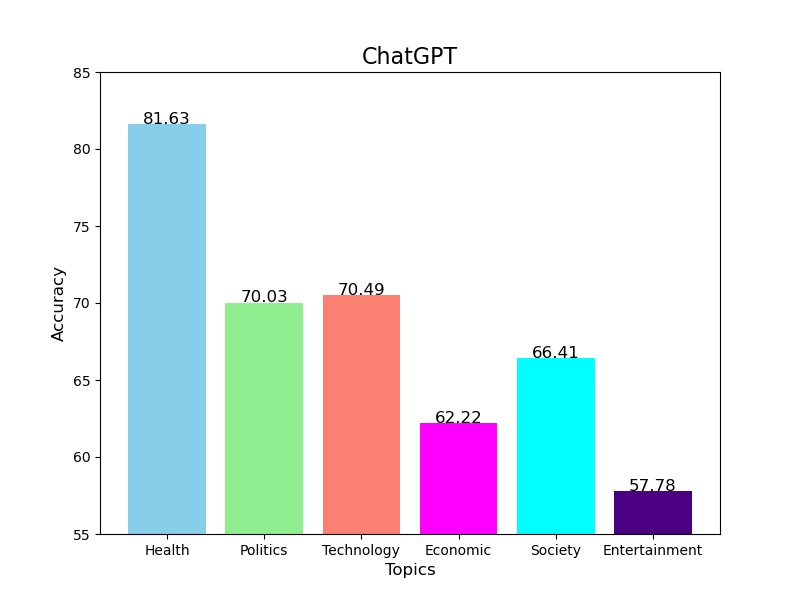}}
  \subfigure[GLM]{\includegraphics[width=0.32\linewidth]{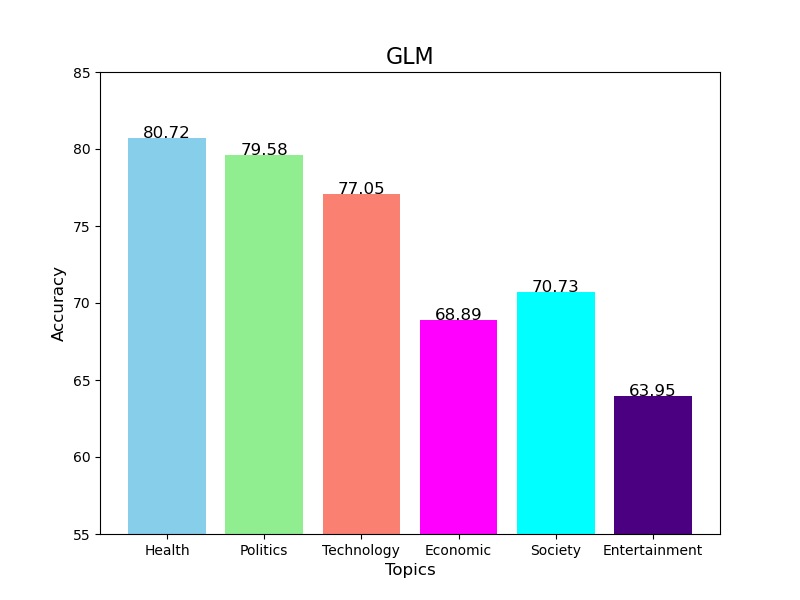}}
    \subfigure[Qwen]{\includegraphics[width=0.32\linewidth]{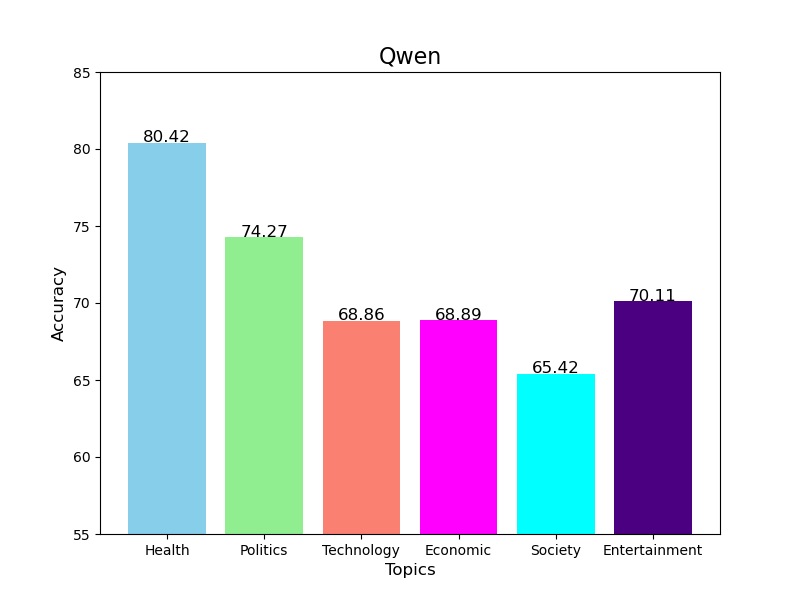}}
  \subfigure[Mistral]{\includegraphics[width=0.32\linewidth]{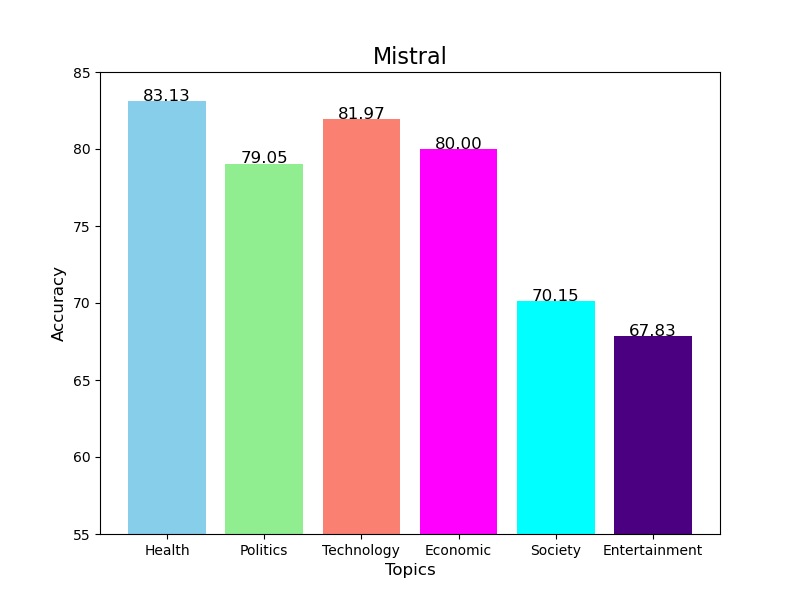}}
  \subfigure[Vicuna]{\includegraphics[width=0.32\linewidth]{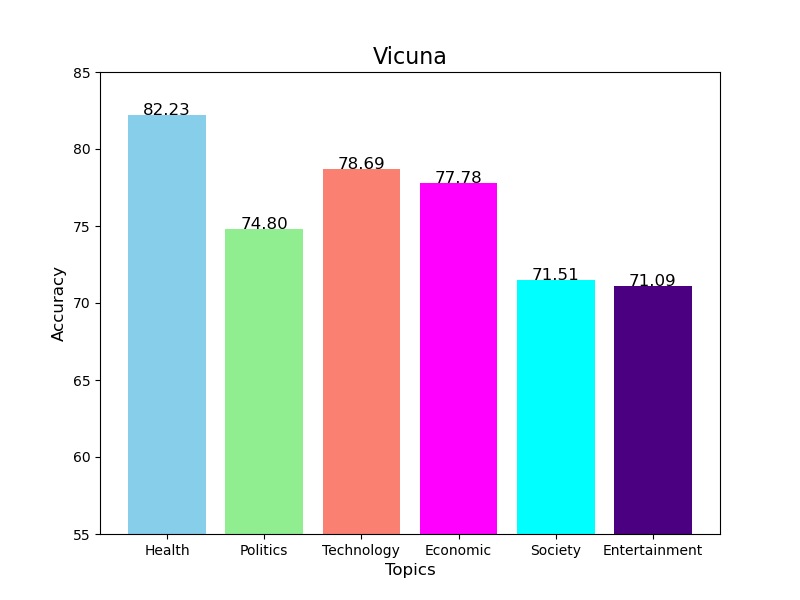}}
  \subfigure[Average performance over LLMs]{\includegraphics[width=0.32\linewidth]{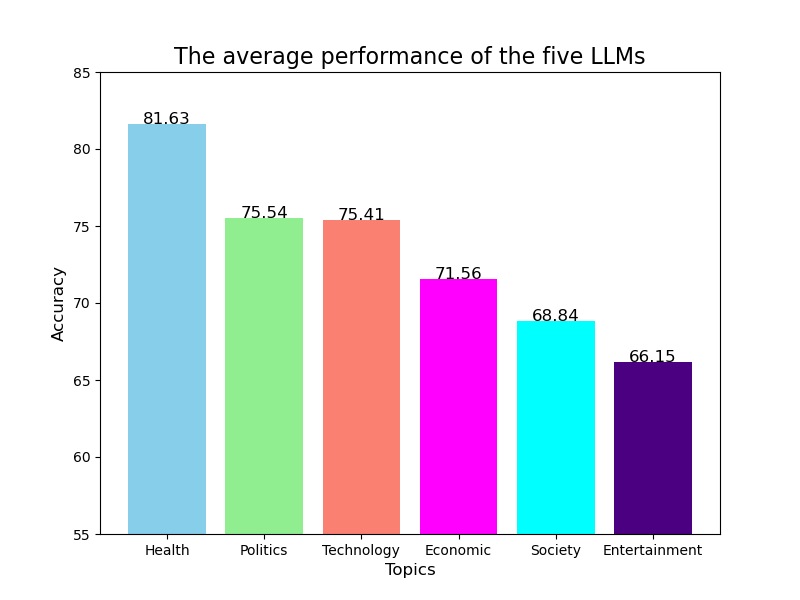}}
\caption{Results (\%) of LLMs against different categories of news.}
  \label{fig:topic}
\end{figure*}

\end{document}